\documentclass[conference]{IEEEtran}
\IEEEoverridecommandlockouts
\usepackage{cite}
\usepackage{amsmath,amssymb,amsfonts}
\usepackage{algorithmic}
\usepackage{graphicx}
\usepackage{textcomp}
\usepackage{xcolor}
\usepackage{soul}
\usepackage{multirow}
\usepackage{comment}

\usepackage{tikz}
\usepackage{textcomp}
\usepackage{hyperref}
\usepackage{lipsum}

\newcommand\copyrighttext{%
  \footnotesize \textcopyright 2024 IEEE. Personal use of this material is permitted.
  Permission from IEEE must be obtained for all other uses, in any current or future 
  media, including reprinting/republishing this material for advertising or promotional 
  purposes, creating new collective works, for resale or redistribution to servers or 
  lists, or reuse of any copyrighted component of this work in other works. 
  DOI: \href{https://ieeexplore.ieee.org/document/10555621}{10.1109/ICHMS59971.2024.10555621}
  }
\newcommand\copyrightnotice{%
\begin{tikzpicture}[remember picture,overlay]
\node[anchor=south,yshift=10pt] at (current page.south) {\fbox{\parbox{\dimexpr\textwidth-\fboxsep-\fboxrule\relax}{\copyrighttext}}};
\end{tikzpicture}%
}

\def\BibTeX{{\rm B\kern-.05em{\sc i\kern-.025em b}\kern-.08em
    T\kern-.1667em\lower.7ex\hbox{E}\kern-.125emX}}
\IEEEoverridecommandlockouts                          
\IEEEpubid{\makebox[\columnwidth]{ 979-8-3503-1579-0/24/\$31.00 ©2024 IEEE  \hfill} \hspace{\columnsep}\makebox[\columnwidth]{ }} 
    
\begin{document}

\title{
Exploring the Dynamics between Cobot's Production Rhythm, Locus of Control and Emotional State in a Collaborative Assembly Scenario
\thanks{This project has received funding from the European Union’s Horizon 2020 research and innovation programme under grant agreement No 847926 MindBot.}
}

\author{
    \IEEEauthorblockN{
        Marta Mondellini$^{1,2}$, Matteo Lavit Nicora$^{1,3}$, Pooja Prajod$^{4}$, Elisabeth André$^{4}$, Rocco Vertechy$^{3}$,
    }
    \IEEEauthorblockN{
        Alessandro Antonietti$^{2}$, Matteo Malosio$^{1}$
    }
    \vspace{1em}
    \IEEEauthorblockA{$^{1}$ CNR, Institute of Intelligent Industrial Technologies and Systems for Advanced Manufacturing, Lecco, Italy.}
    \IEEEauthorblockA{$^{2}$ Catholic University of the Sacred Heart, Department of Psychology, Milan, Italy.}
    \IEEEauthorblockA{$^{3}$ Industrial Engineering Department, University of Bologna, Bologna, Italy.}
    \IEEEauthorblockA{$^{4}$ Human-Centered Artificial Intelligence, University of Augsburg, Augsburg, Germany.}
    \vspace{1em}
    \IEEEauthorblockA{marta.mondellini@stiima.cnr.it}
}

\maketitle
\copyrightnotice

\begin{abstract}
In industrial scenarios, there is widespread use of collaborative robots (cobots), and growing interest is directed at evaluating and measuring the impact of some characteristics of the cobot on the human factor. In the present pilot study, the effect that the production rhythm (C1 - Slow, C2 - Fast, C3 - Adapted to the participant's pace) of a cobot has on the Experiential Locus of Control (ELoC) and the emotional state of 31 participants has been examined. The operators' performance, the degree of basic internal Locus of Control, and the attitude towards the robots were also considered. No difference was found regarding the emotional state and the ELoC in the three conditions, but considering the other psychological variables, a more complex situation emerges. Overall, results seem to indicate a need to consider the person's psychological characteristics to offer a differentiated and optimal interaction experience.

\end{abstract}


\section{Introduction}
Extensive research has been carried out in recent years, focusing on the technical aspects of human-robot interaction (HRI) in industrial scenarios, particularly safety and productivity. However, the experience of a user interacting with this kind of machine is not limited to those aspects and has relevant relapses on a social and psychological level. The need for a careful analysis of how the introduction of autonomous systems influences humans, for example, what effect it has on attention, awareness, and team dynamics, arises. Starting from the analysis of the human factor, a lot of work has been done to obtain indications on which robotic characteristics are socially accepted and make the interaction effective~\cite{koppenborg2017effects, zanchettin2013acceptability, takayama2009influences}. In this sense, recent studies have focused on defining design principles of mental-health-friendly work cells for operators working with cobots~\cite{nicora2021human}. This perfectly aligns with the current need for moving from the concept of Industry 4.0 towards the so-called Industry 5.0~\cite{XU2021530}, that is, the transition from a technology-driven approach towards a value-driven era prioritizing the workers' well-being~\cite{schneiders2022s}. 
In this context, a relevant topic is the identification of which characteristics of a robot influence the emotional state of the subjects who interact with it and other psychological variables, for example, the Locus of Control. 

The "Locus of Control" is the degree to which people believe they have control over events in their lives, rather than being influenced by external forces beyond their influence~\cite{rotter1966generalized}. Thus, it is a one-dimensional construct characterized by two poles, internal and external, placed on the extremities of a continuum; people's attitudes are arranged along this continuum by attributing the cause of what happens to them. 
Individuals with an internal Locus of Control believe events are primarily a result of their actions (e.g., Work performance depends almost entirely on their commitment and abilities). In contrast, those with an external Locus of Control attribute events to external factors (e.g., work performance depends on external factors, including chance). Despite many studies reporting that an internal Locus of Control is more associated with good physical health~\cite{gale2008locus, arraras2002coping, cobb2014healthy, kesavayuth2020locus}, it would seem that even more important is the flexibility with which a person can adapt their thinking (external-internal) depending on the need~\cite{cheng2013cultural}. Regarding robotics, some authors have highlighted that people with an internal Locus of Control have worse usage performance, as they struggle to leave control of the situation to an autonomous system~\cite{takayama2011assisted, acharya2018inference}. Personal experiences, including work experiences, can influence the Locus of Control. 
In particular, a change in the Locus of Control in a specific scenario when interacting/using a product, i.e., a Cobot, is referred to as the Experiential Locus of Control ~\cite{jang2016application}. This concept is an extension of the classic Locus of Control construct and refers to the effect that one experience has on the Locus of Control relative to that specific experience.  

Similarly, for humans and robots to interact effectively, it is necessary to consider the human emotional state in response to robot actions and features, especially in industrial settings, where robots play an important role in automated production~\cite{azni2014analysis}. For example, humans are hypothesized to feel more anxious and cautious when the robot's movement speed is greater~\cite{jafar2014investigation}. One of the aspects to consider, therefore, is the impact that characteristics of the cobot, such as the production rhythm, have on the emotional state. 

 
\section{Aim of The Study}
The present study examined the impact of the cobot's production rhythm on an operator's psychological and emotional state. In particular, how the experiential Locus of Control, the emotional state and performance, and the subjective preference of one experimental condition over another change depending on the cobot's production rhythm was observed. Furthermore, it was observed whether positive or negative attitudes toward robots influence these variables and whether a personal internal Locus of Control impacts the experiential Locus of Control. Further information on these psychological variables and why they are essential in human-robot interaction is reported in the next paragraph.


\section{Materials and Methods}

\subsection{The Experimental Scenario}
\begin{figure}[thpb]
    \centering
    \includegraphics[width=0.50\textwidth]{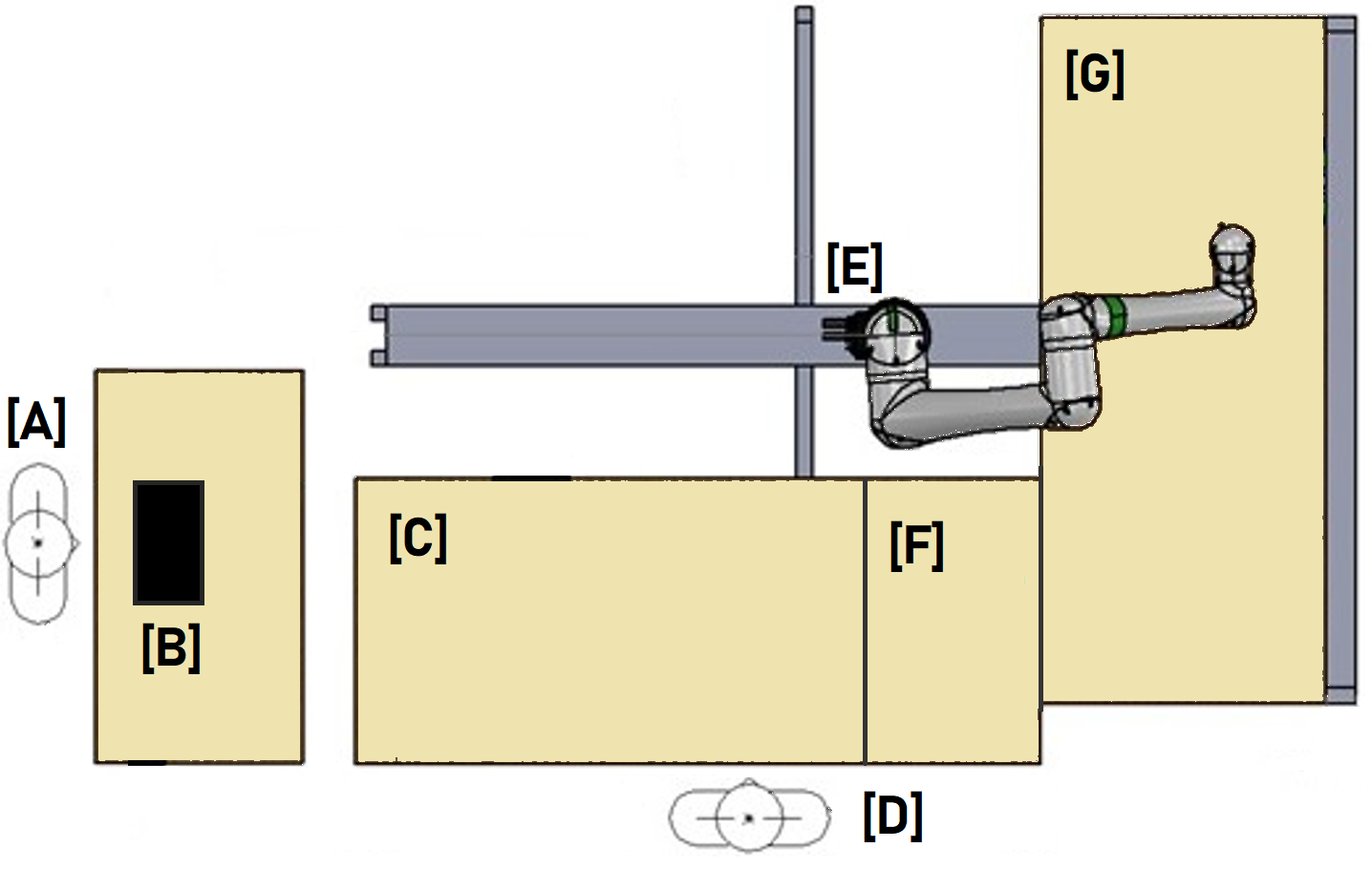}
    \caption{A schematic top view of the work cell where a researcher [A] monitors the session and acts as a Wizard of Oz using a laptop [B]. The participants [D] use the table [C] to work on their part of the assembly, while the cobot [E] moves to the common area [F] for the joint action step after picking up one of the preassembled components available on the other table [G].}
    \label{fig:workcell}
\end{figure}

\begin{figure*}[htbp]
\centerline{\includegraphics[width=0.90\textwidth]{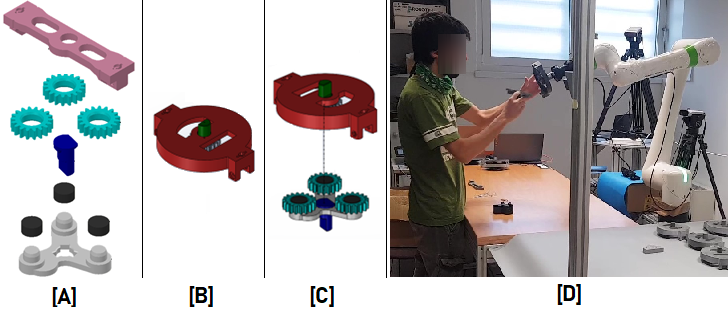}}
\caption{In the picture, [A] shows the components assembled by the user, [B] is the preassembled part assigned to the cobot, [C] shows the joint action step schematically while in [D] on volunteer is depicted in the moment of collaboration with the cobot.}
\label{fig:parts}
\end{figure*}

A collaborative industrial work cell is set up in a lab environment. 
Two distinct areas are defined where the cobot and the operator can work on their sub-assemblies, plus a common area for collaborative joining (Figure~\ref{fig:workcell}). The researcher monitoring the experimental activity uses a dedicated table on the side. This study uses a Fanuc CRX10ia/L cobot equipped with a Robotiq Hand-e parallel gripper to perform the required pick and place operations. The product to be assembled collaboratively by the operator and the cobot is a 3D-printed planetary gearbox~\cite{davide_felice_redaelli_2021_5675810}, represented on the left of Figure~\ref{fig:parts}. Each assembly cycle comprises a sub-assembly phase, followed by a joining phase, during which the two entities work together to mesh the gears and complete the production cycle. The components depicted in Figure~\ref{fig:parts} are assembled by the user, while the pieces assigned to the cobot were already pre-assembled on its side of the table. To control the robot from a custom program running outside the provided teach pendant, a software module is used to interface the robot controller with ROS Noetic~\cite{ros} and, therefore, exploit the capabilities of RosControl~\cite{ros_control} and MoveIt~\cite{coleman2014reducing} tools. On top of that, the high-level state machine needed to orchestrate the task was realized using Visual SceneMaker~\cite{coleman2014reducing}, again interfaced with ROS thanks to dedicated plugins.

\subsection{Study Design}
The study was designed as a within-subjects with three experimental conditions, derived from the experience obtained in~\cite{mondellini14behavioral}. In the mentioned work, a similar setup and task were analyzed, and instances where the cobot or the user would wait for the collaboration to happen, were noticed. Since we expect these instances to lead to different emotions and experiences, we designed the following experimental conditions to explore the topic further:

\begin{enumerate}
    \item\textbf{C1 - Slow condition}: The collaborative robot undergoes a scanning movement across its pre-assembled components, selecting one and delivering it to the user. From the commencement of each production cycle, it typically takes approximately 55 seconds for the cobot to reach the participant for their collaborative task.
    \item\textbf{C2 - Fast condition}: The collaborative robot skips the scanning motion and directly proceeds to the next sub-assembly, picking it up and delivering it to the participant. The cobot takes about 15 seconds from the beginning of each production cycle to reach the participant for their collaborative task.
    \item\textbf{C3 - Adaptive condition}: The collaborative robot executes the scanning motion as described earlier until prompted by the researcher to deliver a specific sub-assembly to the participant. There is no predetermined timing for the cobot, and the cobot is activated through a simple keyboard press whenever the participant is nearing completion of the assembly task.
\end{enumerate}

Each participant performed the three conditions during a single experimental session; presentation order of the conditions was randomized between subjects to control for order effects. Participants were not informed of the cobot's production rhythm change before the experiment.

This study had several objectives, reported below in the form of research questions:
\begin{itemize}
    \item\textbf{Q1}: is the experiential Locus of Control different in the three experimental conditions? 
    \item\textbf{Q2}: is the emotional state different in the three experimental conditions? 
    \item\textbf{Q3}: is the experimental condition possibly influencing performance?
    \item\textbf{Q4}: is the experiential Locus of Control related to the internal Locus of Control?
    \item\textbf{Q5}: is performance related to experiential Locus of Control and emotional state? 
    \item\textbf{Q6}: is the attitude towards robots related to experiential Locus of Control and emotional state?
\end{itemize}

\subsection{Participants}
A total of 33 healthy adult volunteers participated in the experiment (25 males, 8 females) ranging from 18 to 48 y.o. (mean = 29.48, SD = 7.35). The participants were recruited among the institution's employees or students of a nearby University, who were all Italian speakers.
The study has been conducted according to the guidelines of the Declaration of Helsinki and approved by the Ethics Committee of I.R.C.C.S. Eugenio Medea (protocol code N. 19/20-CE of 20 April 2020).

\subsection{Procedure}
The study was carried out as follows. 
The participant entered the room and was asked to read and sign the informed consent. Then, the participant was explained that s/he would have to assemble half the gearbox; s/he was then shown how to do it, and s/he could freely try the assembly until s/he understood the procedure. It was then explained to the participant that the cobot would operate autonomously, bringing the other half of the component into the space used for collaboration; the participant was shown how to operate the joint and finish the product. Although the task was relatively simple and could be completed without the help of the cobot, the subjects were explained that the focus of the study was on the experience of human-machine interaction. The assembly activity would last 15 minutes and be repeated three times. Before the start of the study, and at the end of each 15-minute block, a series of questionnaires designed to evaluate some subjective characteristics and the assembly and interaction experience were administered. The questionnaires were created on the MS form platform and completed online in the same room where the experiment took place.

\subsection{Measures}
The number of assemblies performed during each condition was registered as the participants' performance.

Concerning subjective measures, the following questionnaires were proposed:
\begin{itemize}
    \item\textbf{Internal Control Index}~\cite{duttweiler1984internal} (ICI), administered before the interaction with the cobot. It consists of 28 items to which reply on a 5-point Likert scale, where 1 corresponds to "rarely" and 5 to "usually". A high score (maximum 140) corresponds to a high internal Locus of Control level. The score can vary from 28 to 140.
    \item\textbf{Negative Attitudes Towards Robots Scale}~\cite{nomura2006negative}; this psychometric scale measures negative attitudes towards robots. It consists of 14 questionnaire items divided into three subscales: S1 relates to "negative attitudes toward situations of interaction with robots" (six items), S2 pertains to "negative attitudes toward the social influence of robots" (five items), and S3 addresses "negative attitudes toward emotions in interaction with robots" (three items). Each item is rated on a scale from 1 to 5 (1: strongly disagree - 5: strongly agree). The score can vary from 6 to 30 in S1, 5 to 25 in S2, and 3 to 25 in S3.
    \item\textbf{Experiental Locus of Control Questionnaire} (ELoC), composed of 3 items from the "Internal Control Index"~\cite{duttweiler1984internal}, appropriately modified to evaluate the Experiential Locus of Control~\cite{jang2016application}. The score can vary from 3 to 15.
    \item\textbf{Self-Assessment Manikin}~\cite{bradley1994measuring}; a non-verbal pictorial assessment technique that directly measures the Valence, Arousal, and Dominance associated with a person's emotive reaction to several stimuli. The participant responds on a 9-point Likert scale for each subscale. The Valence scale ranges from positive emotion (happy) to negative emotion (sad). On the Arousal scale, the score varies from high excitement to calmness. Low scores on the Dominance scale correspond to low control, and vice versa. The score can range from 1 to 9. 
\end{itemize}


Additionally, the experimental session ended with a written question on which of the three conditions was preferred. Participants had to justify their choice with a written open answer.

\subsection {Statistical Analysis}
All statistical analyses were performed with SPSS v.29. Skewness and kurtosis have been calculated to evaluate the assumption of normality of the distribution of each variable. Cronbach's alphas were obtained to test the reliability of each subjective scale. Descriptive analyses were run to obtain means and standard deviations for each variable. Repeated measures of ANOVA were performed to compare the scores obtained in the Experiential Locus of Control (Q1) and emotional experience (Q2) in the three experimental conditions. The same test was run to compare the number of assemblies accomplished in the three conditions (Q3). Simple and partial Pearson correlations were run between ICI and the subsequent scores gained on the experiential Locus of Control scale in the three conditions (Q4), between performance, experiential Locus of Control, and emotional state (Q5), and between attitude toward robots and both emotional experience and experiential Locus of Control (Q6). 

\section{Results}

\subsection{Descriptive Statistics}
All the variables considered had normal distribution, except the "performance C3" variable, with kurtosis $>$ 3, and the total score of ICI (kurtosis $>$ 2). Two subjects resulted as outliers in these scales, with Z score $>$ $|$3$|$; after removing them from the database, all variables obtained normal distribution. Therefore, the number of participants whose results were considered was 31 (24 M, 7 F) aged 18 to 48 (mean = 29.61, SD = 7.57).

Descriptive statistics are reported in Table~\ref{tab2}. The reliability of the ICI was very low ($\alpha$ = 0.43); SPSS was asked whether the reliability would be higher by eliminating some items from the scale. A new Internal Control Index was calculated and used in the subsequent inferential analyses ("reduced ICI"); items 7, 8, 17, and 18 of the original questionnaire were deleted. Thus, the lowest score obtainable was 24, and the maximum was 120.

\begin{table}[htbp]
\caption{MEANS, STANDARD DEVIATIONS, MINIMUM AND MAXIMUM SCORES, RANGE
AND ALPHA COEFFICIENTS FOR EACH VARIABLE MEASURED}
\label{tab:my-table}
\begin{tabular}{|l|l|l|l|l|l|}
\hline
\textbf{Variable}                                              & \textbf{Mean} & \textbf{St.Dev.} & \textbf{Min.} & \textbf{Max.} & \textbf{$\alpha$} \\ \hline
\textbf{\begin{tabular}[c]{@{}l@{}}Performance C1\end{tabular}} & 12.90  & 1.11 & 10 & 15 & -    \\ \hline
\textbf{\begin{tabular}[c]{@{}l@{}}Perfomance  C2\end{tabular}} & 17.84  & 2.34 & 13 & 22  &  -    \\ \hline
\textbf{\begin{tabular}[c]{@{}l@{}}Performance C3\end{tabular}} & 16.19  & 1.70 & 12 & 20  & -    \\ \hline
\textbf{ICI}                                                      & 107.48 & 6.47 & 93 & 124 & 0.43 \\ \hline
\textbf{\begin{tabular}[c]{@{}l@{}}reduced\\ ICI\end{tabular}} & 91.68         & 7.00             & 74            & 109            & 0.62       \\ \hline
\textbf{NARS S1}                                                  & 12.03  & 3.43 & 6  & 20   & 0.57 \\ \hline
\textbf{NARS S2}                                                  & 13.03  & 4.10 & 5  & 24   & 0.65 \\ \hline
\textbf{NARS S3}                                                  & 9.45   & 3.60 & 3  & 15   & 0.87 \\ \hline
\textbf{ELoC C1}                                                  & 9.48   & 2.28 & 5  & 13   & 0.50 \\ \hline
\textbf{ELoC C2}                                                  & 9.58   & 2.31 & 5  & 14   & 0.59 \\ \hline
\textbf{ELoC C3}                                                  & 9.52   & 3.01 & 3  & 14  & 0.74 \\ \hline
\textbf{Valence C1}                                               & 4.29   & 1.53 & 1  & 8   & -    \\ \hline
\textbf{Valence C2}                                               & 4.03   & 1.47 & 2  & 8    & -    \\ \hline
\textbf{Valence C3}                                               & 3.84   & 1.46 & 2  & 8    & -    \\ \hline
\textbf{Arousal C1}                                               & 6.84   & 1.66 & 4  & 9    & -    \\ \hline
\textbf{Arousal C2}                                               & 6.35   & 1.76 & 3  & 9    & -    \\ \hline
\textbf{Arousal C3}                                               & 6.77   & 1.91 & 3  & 9   & -    \\ \hline
\textbf{\begin{tabular}[c]{@{}l@{}}Dominance C1\end{tabular}}   & 7.19   & 1.68 & 3  & 9      & -    \\ \hline
\textbf{\begin{tabular}[c]{@{}l@{}}Dominance C2\end{tabular}}   & 7.00   & 1.71 & 4  & 9      & -    \\ \hline
\textbf{\begin{tabular}[c]{@{}l@{}}Dominance C3\end{tabular}}   & 7.03   & 1.70 & 3  & 9       & -    \\ \hline
\end{tabular}
\label{tab2}
\end{table}

\subsection{Differences Between Experimental Conditions}
No difference was observed in the three conditions concerning Experiential Locus of Control(Q1); no difference appeared even when considering emotional Valence, Arousal, and Dominance (Q2).
Regarding the performance, a significant difference was observed between the three conditions, with F(2,60) = 61.68 and p $<$ 0.001. Post-hoc tests with Bonferroni adjustment were run, and differences between conditions 1 and 2 (p $<$ 0.001), between conditions 1 and 3 (p $<$ 0.001), and between 2 and 3 (p = 0.012) emerged (Q3).

\subsection{Correlations Between Variables in Each Condition}
ICI and the Experiential Locus of Control did not correlate in any conditions (Q4). 

Regarding Q5, in C1 some simple and partial correlations emerged, in particular between ELoC and performance (r = 0.36, p = 0.045; r\textsubscript{p} = 0.39, p = 0.38), Valence and Dominance (r = -0.41, p $<$ 0.022), Dominance and Arousal (r = 0.60, p $<$ 0.001; r\textsubscript{p} = 0.61, p $<$ 0.001). In C2 and C3, simple and partial correlation emerged only between Dominance and Arousal: r = 0.65(p $<$ 0.001) and r\textsubscript{p} = 0.66 (p $<$ 0.001) in C2, r = 0.47 (p = 0.007) and r\textsubscript{p} = 0.43 (p = 0.023) in C3.

Concerning Q6, a negative correlation between NARS S1 ("negative attitudes toward situations of interaction with robots") and Valence emerged in C1, with r = -0.43, p = 0.017, and r\textsubscript{p} = -0.52, p = 0.006. The NARS S3 ("negative attitudes toward emotions in interaction with robots") correlated negatively with Dominance, with r = -0.37, p = 0.039 and r\textsubscript{p} = -0.43, p = 0.027.
In C2, only a simple negative correlation between dominance and NARS S1 emerged (r = -0.37, p = 0.045). 
No correlations were found in C3.

\subsection{The Preferred Condition}
Fourteen participants preferred Condition 2 (45.2\%), and 13 participants chose Condition 3 (41.9\%). Only 4 participants (12.9\%) claimed that Condition 1 was the best.

Considering the motivations of these answers, insights emerged. Some participants' replies are reported in Table~\ref{comments}. Those who chose C1 underlined the calmness experienced during the interaction and the feeling of not being under pressure; this result agrees with what is reported on the Arousal scale, where C1 gained scores corresponding to greater calm. Furthermore, some participants said their productivity was the same (even if this was just an impression).
C2 was chosen mainly for its faster pace, and some participants had the feeling that C2 was faster than C3, while others had the opposite impression. In this case, the topic of "competition with her/himself" emerged, and participants reported they felt more engaged in the activity and they worked better. Participants who chose C3 said mainly that the pace was more adequate to their needs, a medium way between C1 and C2; the feeling of being synchronized with the cobot was reported.

\begin{table}[htbp]
\caption{Reasons given for choosing the preferred condition}
\label{comments}
\begin{tabular}{lllll}
\cline{1-2}
\multicolumn{1}{|l|}{\textbf{Favorite Condition}} &
  \multicolumn{1}{l|}{\textbf{Motivation}} &
   &
   &
   \\ \cline{1-2}
\multicolumn{1}{|l|}{\multirow{2}{*}{1}} &
  \multicolumn{1}{l|}{\textit{\begin{tabular}[c]{@{}l@{}}"The robot was slower but productivity\\ did not change"\end{tabular}}} &
   &
   &
   \\ \cline{2-2}
\multicolumn{1}{|l|}{} &
  \multicolumn{1}{l|}{\textit{"I can adapt my pace and work calmly"}} &
   &
   &
   \\ \cline{1-2}
\multicolumn{1}{|l|}{\multirow{3}{*}{2}} &
  \multicolumn{1}{l|}{\textit{"other conditions were too slow"}} &
   &
   &
   \\ \cline{2-2}
\multicolumn{1}{|l|}{} &
  \multicolumn{1}{l|}{\textit{\begin{tabular}[c]{@{}l@{}}"I am a very competitive person \\ and I experienced this condition \\ as a competition"\end{tabular}}} &
   &
   &
   \\ \cline{2-2}
\multicolumn{1}{|l|}{} &
  \multicolumn{1}{l|}{\textit{\begin{tabular}[c]{@{}l@{}}"The robot was faster and didn't slow me \\ down too much. Even though it rushed me\\ a little, it was better than the other two times"\end{tabular}}} &
   &
   &
   \\ \cline{1-2}
\multicolumn{1}{|l|}{\multirow{3}{*}{3}} &
  \multicolumn{1}{l|}{\textit{"the cobot's pace was like mine"}} &
   &
   &
   \\ \cline{2-2}
\multicolumn{1}{|l|}{} &
  \multicolumn{1}{l|}{\textit{\begin{tabular}[c]{@{}l@{}}"the condition had a work pace that \\ was sustainable for a limited time"\end{tabular}}} &
   &
   &
   \\ \cline{2-2}
\multicolumn{1}{|l|}{} &
  \multicolumn{1}{l|}{\textit{\begin{tabular}[c]{@{}l@{}}"it was the condition in which the robotic arm \\ and I were more in symbiosis, I didn't have to \\ wait for it, and it didn't have to wait for me"\end{tabular}}} &
   &
   &
   \\ \cline{1-2}
 &
  
\end{tabular}
\end{table}

\section{Discussion}
The present study examined the influence of the cobot's production rate on participants' experiential Locus of Control, emotion, and performance. 

The production rhythm did not impact ELoC (Q1). No data on this aspect is available in the literature; this is the first result from which to start investigating how and if the production rate of a cobot impacts the worker's locus of control. Regarding Q2, the cobot's production rhythm doesn't affect the participant's emotional state; this result does not agree with other works, where the speed of the cobot increases anxiety~\cite{jafar2014investigation, koppenborg2017effects}. However, in the cited reference, the speed was related to the robot's movement and not only to its production rhythm, which could explain the different results obtained. Additionally, we must consider that the subjects may not have understood the added value of the cobot in this fairly simple assembly task, even though it had been specified to them. The motivational aspect of the activity was, in fact, not considered in this study.

On the other hand, the cobot's production rhythm influenced the participant's performance (Q3), and better performance was accomplished in the condition with the cobot running at the fastest rhythm. This result is understandable, as the slowness of the cobot in C1 forced the participant to wait for the joint activity. However, this result also shows that the subject was pushed to work harder in C2 than s/he would at their own pace, as in C3.

Contrary to what we expected, experiential LoC scores are not correlated with internal LoC (Q4). This could be due to various reasons, but the low validity of the ICI tool that emerged in the statistical analysis suggests the need to investigate this aspect more thoroughly and find alternative tools.

Regardless of the experimental condition, a strong correlation was found between Dominance and Arousal. In short words, a calmer emotional state corresponded to a feeling of greater dominance over the emotion. This result is consistent with what was reported in~\cite{warriner2013norms}. Regarding Q5, a correlation between experiential Locus of Control and performance emerged only in C1; this could indicate that when the subject is not forced to accelerate the pace of production, the Locus of Control of the interaction increases with performance, but forcing the subject to accelerate causes this relationship to disappear. In sum, all correlations between psychological and performance scores are weak. This could be due to the design of the test; for example, the subjects could only start a new assembly after finishing the joint with the cobot; therefore, other variables usually linked to performance, such as the sense of challenge or perceived competence, would not come into play in this correlation.

Concerning Q6, a negative correlation between the negative attitude toward situations of interaction with robots and Valence was observed in C1. This did not happen in C2 and C3, which seems to indicate that setting the cobot to a production rhythm that does not force the subject to speed up leads those with a negative attitude towards interacting with cobots to experience more pleasant feelings. 
Finally, in C1 and C2, a negative correlation between negative attitudes toward emotions in interaction with robots and dominance emerged; this could indicate that in conditions in which the subject's behavior is forced by the speed of the cobot (too slow or too fast), those who tend to have negative emotions while interacting with cobots have less control of their emotional state.

Regarding the favorite condition, C2 and C3 were the most chosen. In the case of C2, the participants appreciated the sense of challenge and the better performance obtained, while in C3, participants perceived a feeling of synchronization with the cobot.

\section{Conclusions, Limitations and Future Work}
This initial investigation offers insights into the design considerations for an industrial setting incorporating a cobot to enhance the operator's well-being. The cobot's production rhythm does not influence the participant's emotional state or Experiential Locus of Control. However, the relationship becomes more multifaceted when the negative attitude toward robots and the personal Locus of Control is considered. In conclusion, these variables, and probably other characteristics such as personality traits, must be regarded as customizing the interaction with the cobot.

The present work has some limitations. Firstly, it is necessary to observe these variables in a larger sample. Furthermore, the "Internal Control Index" did not achieve good reliability, and further studies are needed to confirm the robustness of our findings; for example, the reliability of the scale could be influenced by the sample size or by the Italian language, in which the instrument has not been validated; therefore, it may be helpful to implement an in-depth investigation and modify the questionnaire to make it more robust. 
Similarly, whether this result occurs in samples of different languages should be understood. Additionally, the lack of relationship between ICI and the Experiential Locus of Control suggests that, at least in our sample, the two variables are not linked when, theoretically, they should be. Further observations with a larger sample and more tools dedicated to Locus of Control will need to be conducted to strengthen and confirm our preliminary findings.

Furthermore, a possible limitation of our study is a lack of balance between men and women within the sample. Future studies on these aspects could investigate a potential difference in emotional state and locus of control due to sex.

It would be appropriate to complement these findings with qualitative investigations to explore the conscious motivations that lead a person to prefer one situation over another. An aspect to consider could, for example, be the value the participant gives to the help of the cobot and the motivational aspect in carrying out the task. Additionally, physiological indicators could be associated to gain objective data regarding perceived stress or arousal.


\bibliographystyle{IEEEtran}
\bibliography{biblio}

\end{document}